\documentclass[11pt,letterpaper]{article}
\usepackage{emnlp2016}
\usepackage{times}
\usepackage{latexsym}
\usepackage{amsmath}
\usepackage{multirow}
\usepackage{graphicx}

\usepackage{amsthm}

\emnlpfinalcopy 

\newcommand{\com}[1]{}
\newcommand{\my}[1]{}
\DeclareMathOperator*{\argmax}{arg\,max}

\newcommand{\isection}[2]{\section{#1}\label{ssec:#2}}
\newcommand{\isubsection}[2]{\subsection{#1}\label{ssec:#2}}
\newcommand{\secref}[1]{Section~\ref{ssec:#1}}

\title{Effective Combination of Language and Vision Through Model Composition and the R-CCA Method}

\author{Hagar Loeub \qquad \qquad \qquad \qquad \qquad Roi Reichart \\
	{\tt hagar.loeub@gmail.com} \qquad {\tt roiri@ie.technion.ac.il} \\
	{Faculty of Industrial Engineering and Management, Technion, IIT}
	}

\begin{document}

\maketitle

\begin{abstract}

\end{abstract}

We address the problem of integrating textual and visual information in 
vector space models for word meaning representation. 
We first present the {\it Residual CCA (R-CCA)} method, 
that complements the standard CCA method by representing, for each modality, 
the difference between the original signal and the signal projected to the shared, max correlation, space.
We then show that constructing visual and textual representations and then post-processing 
them through {\it composition of common modeling motifs} such as PCA, CCA, R-CCA and 
linear interpolation (a.k.a {\it sequential modeling}) yields high quality models. 
On five standard semantic benchmarks our sequential models outperform 
recent multimodal representation learning alternatives, including ones that rely on joint representation learning. 
For two of these benchmarks our R-CCA method is part of the {\it Best} configuration our algorithm yields.


\section{Introduction}

\my{1. Compare to Roy; 2. Combination methods following the MIT guy; 
3. read previous work.4. post-processing cascade.} 
\my{TODO: 1. spell check. 2. Dimensionality of visual vectors}

In recent years, vector space models (VSMs), deriving word meaning representations 
from word co-occurrence patterns in text, have become prominent in lexical semantics 
research \cite{Turney:10,Clark:12}. Recent work has demonstrated that when other modalities, 
particularly the visual, are exploited together with text, the resulting multimodal representations 
outperform strong textual models 
on a variety of tasks \cite{Baroni:16}.

Models that integrate text and vision can be 
largely divided to two types. 
{\it Sequential models} first separately construct visual and textual representations and then 
merge them using a variety of techniques: concatenation \cite{Bruni:11,Silberer:13,Kiela:14},
linear weighted combination of vectors 
\cite{Bruni:12,Bruni:14} or linear interpolation of model scores \cite{Bruni:14}, 
Canonical Correlation Analysis and its kernalized version (CCA, \cite{Hill:14,Silberer:12,Silberer:13}), 
Singular Value Decomposition (SVD, \cite{Bruni:14}) 
and Weighted Gram Matrix Combination \cite{Reichart:13,Hill:14}).
{\it Joint models} directly learn a joint representation from textual and visual 
resources using Bayesian modeling \cite{Andrews:09,Feng:10,Roller:13}
and various neural network (NN) techniques: autoencoders \cite{Silberer:14}, 
extensions of word2vec skip-gram \cite{Hill:14a,Lazaridou:15} and others (e.g. \cite{Howell:05}).

The focused contribution of this short paper is two-fold. First, we advocate the {\it sequential approach} for 
text and vision combination and show that when a systematic search in the space of configurations of  
{\it composition of common modeling motifs} \cite{Grosse:14} is employed, this approach 
outperforms recent {\it joint models} as well as sequential models that do not thoroughly search the space 
of configurations. 
This finding has important implications for future research as it advocates the development of efficient search 
techniques in configuration spaces of the type we explore. 

Particularly, we experiment with unimodal dimensionality reduction with Principal 
Component Analysis ((PCA, \cite{Jolliffe:02}), 
multimodal fusion with Canonical Correlation Analysis (CCA, \cite{Hardoon:04}) and 
model score combination with linear interpolation (LI, \cite{Bruni:14}). The composed models 
outperform strong alternatives on semantic benchmarks for word pair similarity 
and association: MEN \cite{Bruni:14}, 
WordSim353 (WS, \cite{Finkelstein:01}), SimLex999 (SL \cite{Hill:15}), SemSim 
and VisSim (SSim, VSim, \cite{Silberer:14}). 

Our second contribution is in proposing the {\it Residual CCA (R-CCA)} method for multimodal fusion. 
This method complements the standard CCA method by representing, for each modality, 
the difference between the original signal and the signal projected to the shared space. 
Since CCA aims to maximize the correlation between the projected signals,
the residual signals intuitively represent uncorrelated components of the original signals. 
Empirically, including R-CCA in the configuration space 
improves results on two evaluation benchmarks. 
Moreover, for all five benchmarks R-CCA substantially outperforms CCA.

\begin{table*}
\scriptsize
\begin{tabular}{| l | l | l | l | l | l | l | l | l | l | l|}
    \hline \% pairs 
    &\multicolumn{2}{|c|}{MEN}  &\multicolumn{2}{|c|}{WS} &\multicolumn{2}{|c|}{SL} &\multicolumn{2}{|c|}{SSim} &\multicolumn{2}{|c|}{VSim}\\
\ (in ImageNet)  &\multicolumn{2}{|c|}{ 42\%} & \multicolumn{2}{|c|}{ 22\%} &\multicolumn{2}{|c|}{ 29\%}  &\multicolumn{2}{|c|}{ 85\%} &\multicolumn{2}{|c|}{ 85\%}\\

	\cline{1-11}
$Model$	& Config. & $\rho$ & Config. & $\rho$ & Config. & $\rho$ & Config. & $\rho$ & Config. & $\rho$ \\
	\hline
	Best &  PCA(200)   & {\bf 0.81} & PCA (250)   & {\bf 0.74} & PCA (250)   & {\bf 0.62} & PCA (50)  & {\bf 0.8} & PCA (100)  & {\bf 0.68} \\
	     &  CCA (V,200)   &            & CCA (250)    &            & CCA (V,250)   &            & CCA (no)  &           & CCA (no) &             \\
         & + R-CCA (T,200) &			&&		&+ R-CCA (T,250)  &		&&		&&		\\
		 &  LI (0.4)  &            & LI  (0.95)  &            & LI  (0.6)  &            & LI  (0.7) &           & LI  (0.45)  &             \\
	\hline Best (No R-CCA)  &  PCA(100)   &			0.79&		&&  PCA(150)   &		0.61&		&& &	 \\
                               &  CCA (50)   &            &    &            & CCA (150)   &            &   &           &&\\
		               &  LI (0.65)  &            &   &            & LI  (0.4)  &            &  &           &  &             \\
    
    \hline 
\hline
    Best (LI) & 0.7 &  0.79 & 0.95 & 0.71 & 0.45 & 0.56& 0.8 & 0.75 & 0.55 & 0.66 \\
	\hline
	Best (PCA, Skip) &  350 & 0.68 &   250 & 0.55 &  300 & 0.54 &  100 & 0.73 &  100 & 0.67\\
	\hline
	Best (PCA, CNN) &  50 & 0.63 &  100 & 0.52 &  300 & 0.53 &  50 & 0.7 &  100  & 0.66 \\
	\hline
	Best (CCA) & V, 100 & 0.59 & T, 150 & 0.61 & V, 50 & 0.47 & T, 50 & 0.39 & V, 100  & 0.35 \\
	\hline
        Best (R-CCA) & L, 50 & 0.77 & L, 150 & 0.71 & V, 300& 0.53 & L, 50 & 0.75 & V, 50 & 0.66 \\
        \hline
	Concatenation & --& 0.63 & -- & 0.48 & -- & 0.54     & -- & 0.57  & -- & 0.6 \\
	\hline	
	\hline
	MMSKIP-A & --& 0.74 & -- & -- & --- & 0.5  & -- & 0.72 & -- & 0.63\\
    \hline
    MMSKIP-B & --& 0.76 & -- & -- & --- & 0.53 & -- & 0.68 & -- & 0.6 \\
	\hline
	BR-EA-14 & --& 0.77 & -- & -- & --- & 0.44  & -- & 0.69 & -- & 0.56\\
	\hline
	KB-14 & --& 0.74 & -- & 0.57 & --- & 0.33  & -- & 0.60 & -- & 0.50\\
	\hline
	\hline
	Skip & --& 0.75 &  -- & 0.73 & -- & 0.46 & -- & 0.73 & -- & 0.67 \\
	\hline
	CNN & -- & 0.58 & -- & 0.54 & -- & 0.53 & --  & 0.67 & -- & 0.64 \\
	\hline
\end{tabular}
\caption{\footnotesize Results. {\textit Best} is the best configuration for each benchmark.
{\it Best (*)} and {\it Concatenation} are the best single-motif models. 
MMSKIP-A and MMSKIP-B are the (joint) models of Lazaridou et al. (2015),  
BR-EA-14 is the best performing model of Bruni et al. (2014) 
and KB-14 is the model of Kiela and Bottou (2014) (both are sequential).
If LI is employed after CCA then its input are the CCA output vectors. If it is employed 
after a CCA+R-CCA then one of its input vectors comes from CCA and the other from R-CCA.
For PCA and CCA we report the number of dimensions and for LI the weight of 
the textual model. For CCA we also report whether the textual (T) or the visual 
(V) projected vector yielded the best result.
{\it Skip} and {\it CNN} are the original input vectors.} 
\label{table:main-results}
\vspace{-0.6cm}
\end{table*}

\vspace{-0.15cm}
\isection {Multimodal Composition}{sec:models}
\vspace{-0.15cm}


\isubsection{Modeling Motifs}{sec:motifs}

\paragraph{PCA}

is a standard dimensionality reduction method. We hence do 
not describe its details here and refer the interested reader to \cite{Jolliffe:02}.


\paragraph {CCA}

finds two projection vectors, one for each original vector, 
such that projecting the original vectors yields the highest possible correlation 
under linear projection.
In short, given an $n$ word vocabulary, with representations $X \in R^{n \times d_1}$ 
and $Y \in R^{n \times d_2}$, CCA seeks two sets of projection vectors 
$V \in R^{d_1 \times d}$ and $W \in R^{d_2 \times d}$ 
that maximize the correlation ($\rho$) between the projected vectors of each of the words:
$V,W = \argmax_{V^{'},W^{'}}  \rho (XV^{'},YW^{'})$. 
The final projection is: $X' = XV$ and $Y' = YW$.

\paragraph {Residual-CCA (R-CCA)}

CCA aims to project the involved representations into a shared space where the correlation between 
them is maximized. The underlying assumption of this method is hence that multiple modalities can facilitate 
learning through exploitation of their shared signal. A complementary point of view would suggest that 
important information can also be found in the dissimilar components of the monomodal signals. 

While there may be many ways to implement this idea, we explore here a simple one which we call the 
{\it residuals approach}. Denoting the original monomodal signals with $X$ and $Y$ 
and their CCA projections with $X'$ and $Y'$ respectively, the {\it residual} signals are defined as:
$R_{x} = X - X'$ and $R_{y} = Y - Y'$. 
Notice that a monomodal signal (e.g. $X$) and its CCA projection (e.g. $X'$) 
may not be of the same dimension. In such cases we first project the original signal ($X$) to 
the dimensionality of the projected signal ($X'$) with PCA.



\paragraph {LI} 

combines the scores produced by two VSMs for a word pair,  
$sc_{m1}(w_i,w_j)$ and  $sc_{m2}(w_i,w_j)$, 
using the linear equation ($\alpha \in [0,1]$): 
{\small $Score (w_i,w_j) = \alpha \cdot sc_{m1}(w_i,w_j) + (1 - \alpha) \cdot sc_{m2}(w_i,w_j)$}.

\isubsection{Motif Composition}{sec:composition}

\begin{figure}[ht]
\centering
\includegraphics [scale=0.4]{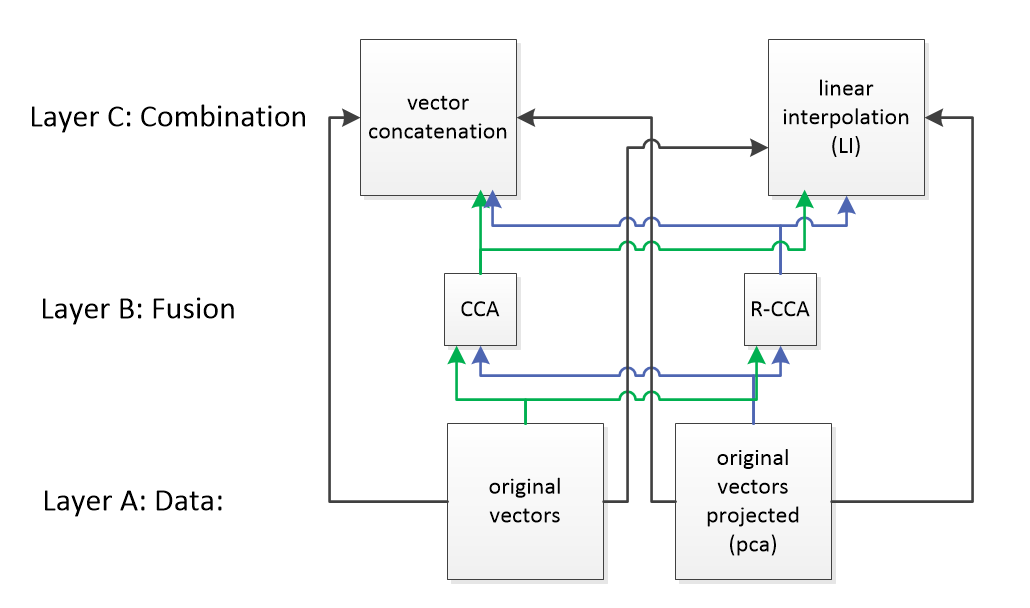}
 \caption{The motif composition search space of the sequential modeling approach for finding
the best composition on pair similarity tasks.}
 \label{motifDiagram}
\end{figure}

We divide the above modeling motifs to three layers, to facilitate 
an efficient systematic optimal configuration search (Figure \ref{motifDiagram}):
(a) Data: (a.1) original vectors; or (a.2) original vectors projected 
with unimodal PCA; 
(b) Fusion: (b.1) CCA and (b.2) R-CCA, each method outputting two projected vectors per word, one for each modality;
(c) Combination: (c.1) vector concatenation; and (c.2) linear interpolation (LI) 
of model scores. 
In our search, a higher layer method considers inputs from all 
lower layer methods as long as both inputs are the output of the same method. 
That is, CCA (layer b.1) is applied to original textual and visual vector pairs 
(output of a.1) as well as to PCA-transformed vectors (a.2), 
but not, e.g., to PCA-transformed visual vectors (a.2) paired with
original textual vectors (a.1).
Vector concatenation (c.1) and linear score interpolation (c.2), in turn, 
are applied to all the inputs and outputs of CCA and of R-CCA. 

The only exception is that we allow an output of CCA (e.g. projected visual vectors) and an output 
of R-CCA (e.g. residual textual vectors) as input to layer c, as two projections 
or two residuals may convey very similar information.
To facilitate efficiency further, CCA and R-CCA are only applied to 
textual and visual vectors of the same dimensionality. 
We leave the exploration of other, possibly more complex, search spaces 
to future work.
\com{\footnote{We also considered a configuration 
where the original vectors are concatenated and then projected with PCA. 
This configuration did not excel on any of our benchmarks.}}

For each benchmark (\secref{sec:experiments}) we search for its {\it {\bf Best}} configuration: 
the optimal sequence of the above motifs, {\bf at most} one from each layer, together with the 
optimal assignment of their parameters.
We do not aim to develop efficient algorithms for
optimal configuration inference, but rather employ an exhaustive grid search approach.
The high quality configurations we find, advocate 
future development of efficient search algorithms.

\isection {Data and Experiments}{sec:experiments}

\paragraph{Input Vectors}

Our textual VSM is word2vec skip-gram \cite{Mikolov:2013c},
\footnote{https://code.google.com/p/word2vec/} 
trained on the 8G words corpus generated by the word2vec 
script.\footnote{\tiny code.google.com/p/word2vec/source/browse/trunk/demo-train-big-model-v1.sh}
We followed the hyperparameter setting of \cite{Schwartz:15} and,  
particularly, set vector dimensionality to 500.
For the visual modality, we used the 
5100 4096-dimensional vectors of \newcite{Lazaridou:15}, 
extracted with a pre-trained Convolutional Neural Network (CNN, \cite{Krizhevsky:12}) 
and the Caffe toolkit \cite{Jia:14} from 100 pictures sampled 
for each word from its ImageNet \cite{Deng:09} entry. 
While there are various alternatives for both textual and visual representations, those we chose are 
based on state-of-the-art techniques.


\paragraph{Benchmarks}

We report the Spearman rank correlation ($\rho$) between 
model and human scores, for the word pairs
in five benchmarks: MEN, 
WS, SL, SSim and VSim. While all the words in our benchmarks appear in our textual corpus, 
only a fraction of them appears in ImageNet, our source of visual input. 
Hence, following  \newcite{Lazaridou:15}, for each benchmark we report results only for  
word pairs consisting of words that are represented in ImageNet.
A model word pair score is  
the cosine similarity between the vectors learned for its words.

\com{
\begin{figure}[t!]
  \centering
    \includegraphics[width=2.5cm,height=2.2cm]{graphs/Simlex_LI.pdf}
    \includegraphics[width=2.5cm,height=2.2cm]{graphs/Simlex_CCA.pdf}
    \includegraphics[width=2.5cm,height=2.2cm]{graphs/Simlex_PCA.pdf}
  \caption{\small Stability analysis for SL. See {\it Stability} paragraph in Section 5 
for details.
The y-axis of all graphs corresponds to Spearman $\rho$. 
For LI, $\alpha$ is the weight of the textual model.}
 \label{fig:graphs}
\vspace{-0.5cm}
 \end{figure}
}

 \paragraph{Parameter Tuning}

We jointly optimized parameters together with the decision of which 
modeling motif to select at each layer, if at all.
For PCA, CCA and R-CCA we iterated over dimensionality values from 50 onward in steps of 50,  
till the minimum dimensionality of the input sets. 
For LI, we iterated over $\alpha \in \{0,0.1,\ldots,1\}$. 
Among the best performing configurations we selected the one 
with the lowest dimension output vectors.
 
Note that there is no agreed upon split of our benchmarks, except from MEN, to 
development and test portions.
Therefore, to facilitate future comparison with our work, 
the main results we report are with the best configuration  
for each benchmark, as tuned on the entire benchmark.  
We also show that our results generalize 
well across evaluation sets, including MEN's dev/test split.  

\paragraph{Alternative Models}

We compare our results to strong alternatives: 
MSKIP-A and MMSKIP-B (\cite{Lazaridou:15}, joint models), the best performing model 
of \newcite{Bruni:14} and \newcite{Kiela:14} (sequential models).\footnote{To facilitate clean
comparison with previous work, 
we copy the results of \newcite{Bruni:14} and of \newcite{Kiela:14} 
from \newcite{Lazaridou:15}, except for WS. 
\newcite{Lazaridou:15} do not report results for WS, while 
\newcite{Bruni:14} report results on a different subset than ours, consisting of 
252 word pairs. \newcite{Kiela:14} report results on our subset of WS, and we copy 
their best result. \newcite{Silberer:14} also report results for SSim and VSim but for 
the entire sets rather than for our subsets.}
While our results are not directly comparable to these models due 
to different training sources and parameter tuning strategies,\footnote{Section 5 
of \newcite{Lazaridou:15} provides the details of the alternative 
models, their training and parameter tuning.} 
this comparison puts them in context.

\isection {Results}{sec:results}



Table \ref{table:main-results} presents the results. 
{\it Best} outperforms the unimodal models
({\textit Skip} and {\textit  CNN}) and the alternative models.
The gains (in $\rho$ points) are: MEN: 4, WS: 1, SL: 9, SSim: 7 and VSim: 1.
R-CCA is included in {\it Best} for MEN and SL, improving over the best
configuration that does not include it by 2 and 1 $\rho$ points, respectively.
Furthermore, R-CCA outperforms CCA on all five benchmarks (by 6-31 $\rho$ points)
and particularly on MEN, SSim and VSim.

\paragraph{Observations}
The five {\it Best} configurations share meaningful patterns. \textbf{(1)} {\it Best} never 
includes concatenation (layer c.1); \textbf{(2)} LI is always included in {\it Best}  
and the weights assigned to the textual and visual modalities are mostly balanced. 
Particularly, the weight of the textual modality is 0.4-0.7 for MEN, SL, SSim and VSim;
\textbf{(3)} In 3 out of 5 cases, {\textit  Best (LI)}, that linearly interpolates the scores yielded by the input 
textual and visual vectors without PCA, CCA or R-CCA processing,
outperforms the models from previous work and the unimodal models; 
\textbf{(4)} In all {\it Best} configurations, the reduced dimensionality is 50-250, 
which is encouraging as processing smaller vectors requires 
less resources. 

\paragraph{Generalization}

We now show that our results generalize well across evaluation sets. 
First, for the portion of the MEN development set that 
overlaps with ImageNet, our {\it Best} MEN configuration 
is the third-best configuration, with $\rho = 0.78$. 
We also tested the {\it Best} configuration as tuned on each of the benchmarks, 
on the remaining four benchmarks. We observed that WS, MEN
and SL serve as good development sets for each other. Best SL 
configuration ({\it Best-SL}): $\rho = 0.7$ on WS and $\rho = 0.77$ on MEN,  
{\it Best-WS}: $\rho = 0.61$ (SL) and $\rho = 0.74$ (MEN), and 
{\it Best-MEN}: $\rho = 0.6$ (SL) and $\rho = 0.71$ (WS). 
The performance of these models on SSim and VSim, however, is substantially lower 
than that of the {\it Best} models of these sets
(e.g. $\rho = 0.62$ for {\it Best-SL} on SSim, $\rho = 0.5$ for {\it Best-WS} on VSim, compared to  
of $\rho = 0.8$ and $\rho = 0.68$, respectively). 
Likewise, SSim and VSim, that consist of the same word pairs scored along different 
dimensions, form good development sets for each 
other ($\rho = 0.66$ for {\it Best-SSim} on VSim, $\rho = 0.78$ for {\it Best-VSim} 
on SSim), but not for WS or SL.
That is, each benchmark has other benchmarks that can serve as its dev. set.

\com{Next, we repeated the following process 5 times: 
for each benchmark we randomly sampled 20\% of the word pairs, and searched 
for the optimal configuration of that subset. 
For SSim and VSim, in all five cases the {\it Best} configuration 
was among the top 3 configurations of the 20\% subset with average 
ranks of 2 and 1.8, respectively. For WS and SL, {\it Best} was among top 4 
configurations in 3 out of 5 cases and for MEN in 1. 
That is, our results for SSim and VSim, but not for the other benchmarks can be reproduced with 
a development(20\%)/test(80\%) split. }

\com{
\paragraph{Stability}

We finally evaluate the stability of the {\it Best} configuration,  
by measuring the change in performance on the entire benchmark, 
where everything is kept fixed except from the parameter of one of the participating methods. 
Figure \ref{fig:graphs} shows the results for SL, the results for the other datasets are very similar. 
For PCA, the resulting $\rho$ scores are stable as long as dimensionality is 500 or lower. 
For CCA, $\rho$ scores change by less than 10 points across dimensionality values, while 
for LI they change by up to $\sim$15 points. This analysis indicates that parameter tuning 
is impactful mostly for CCA and LI, while PCA results are stable across a 
wide range of parameter values.
}

\isection {Conclusions}{sec:conclusions}
 
We demonstrated the power of {\it composition of common modeling motifs} 
in multimodal VSM construction and presented the R-CCA method that exploits the residuals of the CCA signals. 
Our model yields state-of-the-art results on 5 leading semantic benchmarks, for two of which R-CCA  
is part of the {\it Best} configuration. Moreover, R-CCA performs much better than CCA on all five benchmarks.

Our results hence advocate two research directions.
First, they encourage {\it sequential modeling} 
with systematic search in the configuration space for multimodal combination.
Our future goal is making model composition a standard tool for this problem,  
by developing {\it efficient inference algorithms} 
for optimal configurations in possibly more complex search spaces than those 
we explored with an exhaustive grid search. 
Second, the encouraging results of R-CCA emphasize the potential of 
informed post-processing of the CCA output. We intend to deeply delve into this issue in 
the immediate future.
 
\bibliographystyle{emnlp2016}
\bibliography{acl2016}

\end{document}